\pdfoutput=1

\documentclass[11pt]{article}

\usepackage[preprint]{acl}

\usepackage{times}
\usepackage{latexsym}

\usepackage[T1]{fontenc}

\usepackage[utf8]{inputenc}

\usepackage{microtype}
\usepackage{graphicx}
\usepackage{subfigure}
\usepackage{booktabs}

\usepackage{inconsolata}
\usepackage{hyperref}
\usepackage{amsmath}
\usepackage{amssymb}
\usepackage{mathtools}
\usepackage{amsthm}
\usepackage[textsize=tiny]{todonotes}

\usepackage[capitalize,noabbrev]{cleveref}

\theoremstyle{plain}

\theoremstyle{definition}

\theoremstyle{remark}

\usepackage[skins]{tcolorbox} 
\newtcolorbox[auto counter, number within=section, list type=subsubsection, list inside=toc]{sectionbox}[2][]{
	colback=white!98!gray, colframe=black, 
	colbacktitle=white!90!gray, coltitle
	=black, 
	fonttitle=\bfseries,
	title={#2}, 
	list entry={Comment \thetcbcounter\quad}
}
\newcommand{\equalcontrib}{\textsuperscript{\dag}}

\usepackage{graphicx} 
\usepackage{colortbl}
\usepackage{amsmath, amssymb, graphicx}
\usepackage{authblk} 
\usepackage{amsmath,amssymb,amsfonts} 
\usepackage{algorithmic}              
\allowdisplaybreaks[4]               

\title{Multi-Layer GRPO: Enhancing Reasoning and Self-Correction in Large Language Models}

\author{
  \textbf{Fei Ding\equalcontrib},
  \textbf{Baiqiao Wang\equalcontrib},
  
  \textbf{Zijian Zeng\textsuperscript{}},
  \textbf{Youwei Wang\textsuperscript{1}},

\\
  \textsuperscript{1}Shenzhen Jinzheng Youzhi Technology,
\\
  \small{
    \textbf{Correspondence:} \href{mailto:email@domain}{dingfei@email.ncu.edu.cn}}
}

\begin{document}
\maketitle
\begin{abstract}

The Group Relative Policy Optimization (GRPO) algorithm has demonstrated considerable success in enhancing the reasoning capabilities of large language models (LLMs), as evidenced by DeepSeek-R1~\citep{deepseekai2025deepseekr1incentivizingreasoningcapability}. However, the absence of intermediate supervision in GRPO frequently leads to inefficient exploration dynamics. A single error in a complex reasoning chain can invalidate the entire solution, resulting in abrupt reward vanishing and compromising training stability.To address these challenges, we propose MGRPO (Multi-layer GRPO). MGRPO operates in two layers: the first layer employs standard GRPO to generate an initial response. This response, along with the original query, is then fed into a second-layer GRPO process. This second layer is specifically trained to identify and correct errors in the initial response, effectively creating a self-correction loop.
This mechanism provides implicit process-level supervision by rewarding successful error correction, without requiring an explicit, densely-annotated reward model. Experimental results on several mathematical reasoning benchmarks demonstrate that MGRPO significantly outperforms standard GRPO, achieving superior performance by fostering both reasoning and self-correction abilities.

\end{abstract}

\section{Introduction}

The rule-based GRPO framework has demonstrated remarkable efficacy in the DeepSeek-R1 implementation.However, pure Sparse Outcome Rewards provide feedback only at the final step of reasoning, which often leads to inefficiencies in reinforcement learning (RL) for large language models (LLMs). A minor mistake in an intermediate step can result in a completely incorrect final answer, leading to zero rewards and making the exploration process highly inefficient~\citep{uesato2022solving,Lightman2023LetsVS,Wang2023MathShepherdVA,yuan2024freeprocessrewardsprocess}.

In contrast, Dense Process Rewards demonstrate significant advantages in terms of reasoning scalability, training efficiency~\citep{sutton2018reinforcement}, and credit assignment~\citep{leike2018scalable}. However, the practical application of dense rewards in RL training remains limited~\citep{setlur2024rewarding}, and current state-of-the-art industrial models still primarily rely on verifiable outcome rewards without achieving breakthroughs in dense reward utilization~\citep{deepseekai2025deepseekr1incentivizingreasoningcapability,team2025kimi}. The core challenge lies in how to efficiently obtain and leverage high-quality dense rewards~\citep{Gao2022ScalingLF}.Dense rewards are mainly obtained by reward models, while training a exrta reward model poses great challenges,sucn as more resources consumption and how to guarantee the accuracy of rewards.

To address these challenges, we propose Multi-Layer GRPO (MGRPO). This approach introduces a form of implicit process self-supervision on top of outcome-based rewards, effectively unifying the generative model and a process-aware refinement mechanism within a single, cohesive framework. MGRPO employs a two-layer GRPO process for each data instance:

\textbf{First Layer(Standard GRPO)}: The model generates initial responses using GRPO, following the conventional approach.

\textbf{Second Layer}: The input and process output from the first stage are concatenated, followed by some guiding prompts as input, guiding the model to identify and correct errors. If errors cannot be corrected, the data is discarded. If successfully revised, GRPO training is applied to reinforce the model’s ability to rectify mistakes.This process implicitly includes a reward for the first layer GRPO process.

By leveraging the model's own output as a basis for critical review and correction, MGRPO encourages the model to learn from its errors. This method achieves an effect similar to dense process rewards without requiring an external reward model to provide explicit intermediate reward values. Experimental results demonstrate that MGRPO achieves superior performance compared to standard GRPO alone by cultivating both primary reasoning and self-correction skills.

Overall, our contibutions are as follows:
\begin{itemize}
	\item We propose \textbf{MGRPO}, a structured multi-layer learning paradigm that enables LLMs to simultaneously enhance their reasoning capabilities and error correction abilities, allowing the model to learn from its own mistakes.
	\item We introduce a \textbf{correction-augmentation-selection mechanism} for the second-layer GRPO. This involves selectively focusing on successfully corrected responses and potentially augmenting the training data with these successful correction trajectories, ensuring more robust and efficient learning of self-correction skills.
	\item Extensive experiments on multiple mathematical reasoning benchmarks demonstrate that MGRPO significantly outperforms standard GRPO and other relevant baselines.
\end{itemize}

\section{Related Work}

\subsection{GRPO}
The Group Relative Policy Optimization (GRPO) algorithm, as prominently featured in the DeepSeek-R1 model~\citep{deepseekai2025deepseekr1incentivizingreasoningcapability}, has marked a significant advancement in applying reinforcement learning to enhance LLM reasoning. GRPO distinguishes itself by forgoing an explicit value function, instead using the average reward of a group of sampled outputs for the same query as a baseline for advantage estimation. This design aligns well with rule-based reward systems where final outcomes (e.g., correctness of a math solution) can be programmatically verified. 

Compared with PPO~\citep{schulman2017proximal}, GRPO achieves substantial reductions in computational resource consumption and marked improvements in training effectiveness. Nevertheless, GRPO frequently faces challenges like entropy collapse, reward collapse, and training instability~\citep{yu2025dapo} during the training process due to the lack of process supervision. 

\subsection{Self-Correction Fine-Turning}
The idea of enabling LLMs to correct their own mistakes has gained considerable traction and been demonstrated effective to improve LLMs' accuracy on intricate assignments.

In the SFT paradigm, \citet{li-etal-2024-selective} presented a method where a teacher model generates multiple reflective responses, and a student model is fine-tuned to select the most compatible one, thereby enhancing the student's self-reflection capability. \citet{huang2023large} explored "intrinsic self-correction" by prompting models to review and correct their outputs, showing some capability but also limitations, particularly in knowing when to stop or how to effectively correct.

 \citet{kumar2024training} proposed SCoRe (Self-Correction via Reinforcement Learning), which trains LLMs using self-generated data through multiple rounds of online RL. SCoRe specifically addresses challenges like distribution mismatch (training on errors the model itself is likely to make) and behavioral collapse (where the model becomes too good at the initial generation, leaving no room for correction). While effective, SCoRe involves multi-stage RL and reward shaping, adding much complexity.

\section{Methods}

\subsection{Preliminaries: Group Relative Policy Optimization (GRPO)}
We first briefly introduce the Group Relative Policy Optimization (GRPO) algorithm \citep{shao2024deepseekmath, deepseekai2025deepseekr1incentivizingreasoningcapability}, which forms the basis for each layer in our proposed MGRPO framework. 

Given an input query $q$, GRPO samples a group of $G$ responses $\{o_1, o_2, \dots, o_G\}$ from the current policy $\pi_{\theta_{old}}$. The core idea is to update the policy $\pi_\theta$ by maximizing an objective function that encourages responses with higher-than-average rewards within their group. The GRPO objective function is defined as:
\begin{equation}
\resizebox{\linewidth}{!}{$\displaystyle
    \begin{aligned}
        \mathcal{J}_{\text{GRPO}}(\theta) 
        &= \mathbb{E}_{\substack{q \sim P(Q),  \{o_i\}_{i=1}^G \sim \pi_{\theta_{old}}(O|q)}} 
        \Bigg[ \frac{1}{G} \sum_{i=1}^G \frac{1}{|o_i|} \sum_{t=1}^{|o_i|} \\
        &\quad \min \bigg( 
            \frac{\pi_\theta(o_{i,t} | q, o_{i,<t})}{\pi_{\theta_{old}}(o_{i,t} | q, o_{i,<t})} \hat{A}_{i,t}, \\
        &\quad \text{clip} \left( 
            \frac{\pi_\theta(o_{i,t} | q, o_{i,<t})}{\pi_{\theta_{old}}(o_{i,t} | q, o_{i,<t})}, 1 - \epsilon, 1 + \epsilon 
        \right) \hat{A}_{i,t} 
        \bigg) \\
        &\quad - \beta \mathbb{D}_{KL} \left[ \pi_{\theta} \parallel \pi_{ref} \right] \Bigg],
    \end{aligned}$}
    \label{eq:GRPO-obj}
\end{equation}
where: 

-$o_{i,t}$ is the $t$-th token of the $i$-th response $o_i$.
 
-$\frac{\pi_\theta(o_{i,t} | q, o_{i,<t})}{\pi_{\theta_{old}}(o_{i,t} | q, o_{i,<t})}$ is the importance sampling ratio for token $o_{i,t}$.

-$\hat{A}_{i,t}$ is the advantage estimate for token $o_{i,t}$ of response $o_i$. In GRPO, this advantage is typically calculated as $R(o_i) - \bar{R}$, where $R(o_i)$ is the total reward for response $o_i$, and $\bar{R} = \frac{1}{G} \sum_{j=1}^G R(o_j)$ is the average reward of the group of $G$ sampled responses. The same advantage $\hat{A}_{i} = R(o_i) - \bar{R}$ is often applied to all tokens $t$ within a given response $o_i$.
 
-$\epsilon$ is a hyperparameter for clipping the importance ratio, similar to PPO, to stabilize training.
 
-$\beta$ is a hyperparameter controlling the strength of the KL divergence penalty.
 
-$\pi_{ref}$ is a reference policy (often the initial supervised fine-tuned model) used to prevent the learned policy $\pi_\theta$ from deviating too far.
 
 The KL divergence $\mathbb{D}_{KL}[\pi_{\theta} || \pi_{ref}]$ is estimated using an unbiased estimator proposed by \citet{kl_approx}:

\begin{equation}
\resizebox{0.8\linewidth}{!}{$\displaystyle
	\mathbb{D}_{KL}\left[\pi_{\theta} || \pi_{ref}\right] \approx \mathbb{E}_{o \sim \pi_\theta} \left[ \log\frac{\pi_{\theta}(o|q)}{\pi_{ref}(o|q)} \right]$}
\end{equation}
or more practically, often approximated token-wise during optimization as:
\begin{equation}
\resizebox{\linewidth}{!}{$\displaystyle
\begin{aligned}
    \mathbb{D}_{KL}\left[\pi_{\theta}(o_{i,t}|q,o_{i,<t}) || \pi_{ref}(o_{i,t}|q,o_{i,<t})\right] \approx \\ \frac{\pi_{ref}(o_{i,t}|q,o_{i,<t})}{\pi_{\theta}(o_{i,t}|q,o_{i,<t})} - \log\frac{\pi_{ref}(o_{i,t}|q,o_{i,<t})}{\pi_{\theta}(o_{i,t}|q,o_{i,<t})} - 1,
\end{aligned}$}
\label{eq:kl_approx_tokenwise}
\end{equation}
This KL term ensures that the policy updates are not too drastic, maintaining stability.

\subsection{MGRPO: Multi-Layer Group Relative Policy Optimization}
MGRPO extends the standard GRPO framework by introducing a two-layer hierarchical structure. This structure is designed to first generate an initial solution and then to explicitly train the model to review and correct this solution, thereby fostering both reasoning and self-correction abilities.

\subsubsection{Overall MGRPO Process}
For each data instance (query $q$):
\begin{enumerate}
    \item \textbf{Layer 1 (Initial Response Generation):} The model, using its current policy $\pi_\theta$, generates a set of $G$ initial responses $\{o_1, o_2, \dots, o_G\}$ to the query $q$. These responses are evaluated (e.g., by a rule-based verifier for correctness), and the GRPO objective (Equation \ref{eq:GRPO-obj}) is used to update the policy based on these initial attempts.
    \item \textbf{Layer 2 (Self-Correction and Refinement):} Each initial response $o_i$ from Layer 1, along with the original query $q$, is used to form a new, augmented query $q'_i$ for the second layer. The model then generates a set of $H$ "corrected" or "refined" responses $\{\tilde{o}_{i,1}, \dots, \tilde{o}_{i,H}\}$ for each $q'_i$. These refined responses are also evaluated. The GRPO objective is again applied, this time to update the policy based on its performance in the self-correction task.
\end{enumerate}
The policy $\pi_\theta$ is shared across both layers, meaning the same model parameters are updated by gradients from both the initial problem-solving phase and the subsequent self-correction phase.

\subsubsection{Prompt Engineering for MGRPO}
Effective prompting is crucial for guiding the LLM's behavior in both layers. We employ a concise template to guide the base model in adhering to specified instructions. As shown in Table \ref{tab:r0_template}, this template mandates thorough reasoning before providing an answer. We intentionally constrain this structured format to facilitate rule-based reward modeling.
This prompt is based on the DeepSeek-R1~\citep{deepseekai2025deepseekr1incentivizingreasoningcapability} paper. We found that there are some issues with its prompt, specifically that the terms "answer" and the \texttt{<answer></answer>} tags are incorrect. This causes the model to provide only the final answer within the \texttt{<answer></answer>} tags, while omitting the reasoning process.Therefore, we have made some improvements.

\begin{table*}[t]
	\centering
	\begin{minipage}{2.0\columnwidth}\vspace{0mm}    \centering
	    \caption{Template for the initial query in MGRPO (Layer 1). \textcolor{red}{[Original Problem Prompt]} will be replaced with the specific reasoning question during training.
		} 
		\label{tab:r0_template}
		\begin{sectionbox}[]
			\centering
			\small
			\begin{tabular}{p{0.97\columnwidth} c}

			   \texttt{<|im\_start|>system}
			   
			   \texttt{You are a helpful AI assistant. A conversation between User and Assistant.}
			   
			   \texttt{The User asks a mathematical question, and the Assistant solves it step-by-step.}
			   
			   \texttt{The Assistant must first output a detailed step-by-step reasoning process enclosed within <think></think> tags. After the </think> tag, the Assistant must provide the final answer based on the reasoning.}
			   
			   	\texttt{<|im\_end|>}
			   	
			   	\texttt{<|im\_start|>user}
			   	
			   	\textcolor{red}{[Original Problem Prompt]}
			   			
			   			\texttt{<|im\_end|>}
			   				   	
			   	\texttt{<|im\_start|>assistant}
				\\\\

			\end{tabular}
		\end{sectionbox}
	\end{minipage}
\end{table*}

For Layer 2, the input prompt is constructed to encourage self-reflection and correction. This involves concatenating the original query, the model's Layer 1 output, and a specific guiding phrase.

\subsubsection{Layer 1 GRPO: Initial Response Generation}
In the first layer, MGRPO operates identically to standard GRPO. Given an input query $q$, the model $\pi_{\theta_{old}}$ generates $G$ responses $\{o_1, \dots, o_G\}$. Each response $o_i$ consists of a reasoning trace and a final answer. A rule-based verifier assigns a reward $R(o_i)$ . The policy $\pi_\theta$ is then updated using the GRPO objective:
\begin{equation}
\resizebox{\linewidth}{!}{$\displaystyle 
	\begin{aligned}
		\mathcal{J}_{\text{Layer1-GRPO}}(\theta) = & \mathbb{E}_{q \sim P(Q), \{o_i\}_{i=1}^G \sim \pi_{\theta_{old}}(O|q)}
		\Bigg[ \frac{1}{G} 
		\sum_{i=1}^{G}\\ \frac{1}{|o_i|} \sum_{t=1}^{|o_i|}  
		& \min \bigg( r_{i,t}^{(L1)}(\theta) \hat{A}_{i,t}^{(L1)}, \\
		& \quad \text{clip}(r_{i,t}^{(L1)}(\theta), 1 - \epsilon, 1 + \epsilon)  \hat{A}_{i,t}^{(L1)} \bigg) \\
		& - \beta \mathbb{D}_{KL}(\pi_{\theta} || \pi_{ref}) \Bigg],
	\end{aligned}$}
    \label{eq:Layer1-GRPO-obj}
\end{equation}
where $r_{i,t}^{(L1)}(\theta) = \frac{\pi_\theta(o_{i,t} | q, o_{i,<t})}{\pi_{\theta_{old}}(o_{i,t} | q, o_{i,<t})}$ is the importance ratio for Layer 1, and $\hat{A}_{i,t}^{(L1)}$ is the advantage calculated based on the rewards of the $G$ initial responses for query $q$. This layer primarily trains the model's ability to solve the problem directly.

\subsubsection{Layer 2 GRPO: Self-Correction and Refinement}
The second layer is the core of MGRPO's self-correction mechanism. It involves a correction-augmentation-selection mechanism. 

\textbf{Correction:}For each initial response $o_i$ generated in Layer 1 (containing the thought process $\text{thought}_i$ and answer $\text{ans}_i$), a new input $q'_i$ for Layer 2 is constructed as:
\begin{equation}
\resizebox{0.8\linewidth}{!}{$\displaystyle 
   \begin{aligned}
    q'_i = \text{SystemPrompt}_{\text{L2}} + \text{User:} [q] + \text{Assistant:}\\ [\text{\textless think\textgreater}\text{thought}_i\text{\textless/think\textgreater}\text{ans}_i] + p_{\text{guide}.}
    \end{aligned}$}
    \label{eq:layer2_input_formulation}
\end{equation}
$\text{SystemPrompt}_{\text{L2}}$ is a system prompt tailored for the correction task. 
$[q]$ is the original problem query.
$[\text{{\textless
think\textgreater}thought}_i\text{{\textless/think\textgreater}ans}_i]$ is the full output from Layer 1 for the $i$-th sample.
$p_{\text{guide}}$ is a guiding phrase randomly selected from a predefined pool to prevent the model from overfitting to a single template. Examples are as follows:  
	\begin{itemize}
		\item Where might I have gone wrong this time? Let me double-check carefully.
		\item Wait, let me double-check that.
		\item Wait a minute, let me make sure I didn't make a mistake.
		\item Hmm, let me think if there's another way to approach this problem.
		\item Wait, maybe I can think about it like this:
		\item Another thought: maybe I can 
		\item But wait, let me just make sure I didn't miss anything in the original problem.
	\end{itemize}

\textbf{Augmentation:}For each $q'_i$, the model $\pi_{\theta_{old}}$ generates $H$ candidate corrected responses $\{\tilde{o}_{i,1}, \dots, \tilde{o}_{i,H}\}$. The parameter $H$ represents the number of correction attempts (or augmented variations) sampled for each $q'_i$. By sampling multiple ($H > 1$) such responses, we gather more data for how the model attempts to correct, which is particularly valuable for successful corrections, strengthening the learning signal for effective refinement strategies.

\textbf{Selection:} Each corrected response $\tilde{o}_{i,j}$ is evaluated by the rule-based verifier, yielding a reward $R(\tilde{o}_{i,j})$. A crucial step here is dynamic verification:
        \begin{itemize}
            \item If the Layer 1 response $o_i$ was incorrect, and a Layer 2 response $\tilde{o}_{i,j}$ is correct, this is a successful correction.
            \item If $o_i$ was correct, and $\tilde{o}_{i,j}$ remains correct (ideally with similar or improved reasoning, though current verification is outcome-based), this is a successful confirmation.
            \item If $o_i$ was correct but $\tilde{o}_{i,j}$ becomes incorrect, or if $o_i$ was incorrect and $\tilde{o}_{i,j}$ remains incorrect or worsens, these are unsuccessful correction attempts.
        \end{itemize}
        
For MGRPO training, we primarily focus on instances where the Layer 2 process leads to a correct final answer. Trajectories where $o_i$ was incorrect and all $H$ corrected responses $\tilde{o}_{i,j}$ remain incorrect would be discarded, not contributing to the Layer 2 gradient update.

The policy $\pi_\theta$ is then updated using a GRPO objective similar to Layer 1, but based on the corrected responses and their rewards. \begin{equation}
\resizebox{\linewidth}{!}{$\displaystyle
	\begin{aligned}
		\mathcal{J}_{\text{Layer2-GRPO}}(\theta) = & \mathbb{E}_{q' \sim P(Q'), \{\tilde{o}_j^{(h)}\}_{h=1}^{H} \sim \pi_{\theta_{old}}(O|q')} \\
		& \Bigg[ \frac{1}{H} \sum_{j=1}^{G'} \sum_{h=1}^{H} \frac{1}{|\tilde{o}_j^{(h)}|} \sum_{t=1}^{|\tilde{o}_j^{(h)}|}  \\
		& \min \Big( \tilde{r}_{j,t}^{(h)}(\theta) \hat{A}_{j,t}^{(h)}, \\
		& \quad \text{clip}(\tilde{r}_{j,t}^{(h)}(\theta), 1 - \epsilon, 1 + \epsilon)  \hat{A}_{j,t}^{(h)} \Big) \\
		& - \beta \mathbb{D}_{KL}(\pi_{\theta} || \pi_{ref}) \Bigg].
	\end{aligned}$}
\end{equation}
where:

- \( q' \) is the modified query after correction.

- \( \tilde{o}_j^{(h)} \) represents the corrected, selected, and augmented responses.

- \( \tilde{r}_{j,t}^{(h)}(\theta) \) and \( \hat{A}_{j,t}^{(h)} \) are the importance ratio and relative advantage estimates for Layer2.

This two-layer process enables the model to not only learn how to solve problems (Layer 1) but also how to identify and fix its own errors (Layer 2). The successful correction in Layer 2 provides an implicit positive reward signal for the intermediate reasoning steps, guiding the model towards more robust and accurate problem-solving strategies.

\section{Experiment}
\label{sec:experiments}

\paragraph{Tasks} We evaluate MGRPO primarily on challenging mathematical reasoning datasets:
\begin{itemize}
    \item \textbf{MATH}~\citep{hendrycksmath2021}: A dataset of 12,500 challenging competition mathematics problems. We report results on the remaining 500 problems from the test set, referred to as MATH500, for evaluation.
    \item \textbf{GSM8K}~\citep{cobbe2021training}: A dataset of 8,500 grade school math word problems. We report results on its 1000 test samples.
    \item \textbf{Minerva Math}~\citep{lewkowycz2022solving}: This benchmark comprises STEM problems at the undergraduate level, requiring quantitative reasoning across various scientific and mathematical domains. It is designed to assess multi-step reasoning capabilities.
    \item \textbf{OlympiadBench}~\citep{he2024olympiadbench}: A collection of 8,476 advanced math and physics problems from international and national Olympiads and college entrance exams, each with expert-level, step-by-step solutions. This benchmark tests high-level reasoning and problem-solving.
\end{itemize}
For all tasks, the final answer is extracted from the model's generation and compared against the ground truth. A rule-based verifier checks for numerical equivalence for math problems.

\paragraph{Base Model and Inference}
The base model for all our experiments is \textbf{Qwen2.5-Math-7B-base}. This model is specifically pre-trained for mathematical reasoning, providing a strong foundation.

\begin{table*}[t]
	\centering 
	\caption{Performance comparison on mathematical reasoning benchmarks using Qwen2.5-Math-7B-base. Temperature set to 0.7 for generation. One-round baselines (PPO, GRPO) do not have an explicit self-correction turn; Acc.@t1 is their final accuracy. For MGRPO, Acc.@t1 is the Layer 1 output accuracy, Acc.@t1$^\prime$ is the accuracy after the Layer 2 correction attempt (before Layer 2 GRPO update parameters ), and Acc.@t2 is the final accuracy after the full MGRPO process. For Intrinsic self-correction, Acc.@t1$^\prime$ denotes the accuracy after the initial correction prompt, while Acc.@t2 represents the accuracy achieved following two rounds of self-correction.}  
	\label{tab:scaled_result}
	\renewcommand{\arraystretch}{1.1}
	\vspace{0cm}
	\resizebox{1.0\textwidth}{!}{
		\begin{tabular}{c|c|ccc|c|cc}
			\toprule
			\textbf{Benchmark} & \textbf{Method}  & \textbf{Acc.@$t_1$} & \textbf{Acc.@$t_1^\prime$} & \textbf{Acc.@$t_2$} & $\Delta(t_1^\prime,t_2)$ & $\Delta^{i \to c}(t_1^\prime,t_2)$ & $\Delta^{c \to i}(t_1^\prime,t_2)$ \\ 
			\midrule 
			\midrule 
			&   One-round PPO               & 79.3 & - & - & - & - & - \\
			MATH &   One-round GRPO              & 80.9 & - & - & - & - & - \\
			& Intrinsic self-correction  & - & 65.1  & 52.7  & -12.4 & 2.6 & 15.0 \\
			\rowcolor[rgb]{ .867, .922, .169} &MGRPO & 80.9 & 87.5 & 90.4 & +2.9 & 3.0 & 0.1 \\
			\midrule 
			\midrule 
			&   One-round PPO               & 82.2 & -& - & - & - & - \\
			GSM8K &   One-round GRPO              & 83.4 & -& - & - & - & - \\
			& Intrinsic self-correction & - & 77.9 & 71.1 & -6.8 & 5.1 & 11.9 \\
			\rowcolor[rgb]{ .867, .922, .169} &MGRPO & 83.4& 93.4 & 95.6 & +2.2 & 2.3 & 0.1 \\
			\midrule 
			\midrule
			& One-round PPO & 33.3 & - & - & -& - & - \\
			Minerva Math & One-round GRPO & 35.1 & -& - & - & - & - \\
			&Intrinsic self-correction & -& 21.3 & 17.8  & -3.5 & 2.3 & 5.8 \\
			\rowcolor[rgb]{ .867, .922, .169} &MGRPO & 35.1 & 36.1 & 39.3 & +3.2 & 3.9 & 0.7\\ 
			\midrule 
			\midrule
			& One-round PPO & 39.8 & -& - & - & - & - \\
			OlympiadBench& One-round GRPO & 39.9 & -& - & - & - & - \\
			& Intrinsic self-correction & -& 27.3 & 22.4 & -4.9 & 2.5 & 7.4 \\
			\rowcolor[rgb]{ .867, .922, .169} &MGRPO & 39.9 & 45.5 & 50.4 & +4.9 & 5.3 & 0.4\\
			\bottomrule
		\end{tabular}
	}
    \vspace{10pt}
\end{table*}

\paragraph{Evaluation Metrics}
To comprehensively assess the model's performance, particularly its self-correction capabilities, we adopt the following metrics, inspired by \citet{kumar2024training}:
\begin{enumerate}
	\setlength{\itemsep}{1pt}
	\item \textbf{Accuracy@t1}: accuracy of the first turn RL;
	\item \textbf{Acc.@t1$^\prime$}: accuracy after the first turn RL and self-correction under the guidance of prompts (If the answer is correct multiple times, half of the points will be awarded.);
	\item \textbf{Accuracy@t2}: accuracy of the second turn RL;
	\item $\Delta(t_1^\prime,t_2)$: improvement in accuracy due to the second turn RL;
	\item $\Delta^{i \to c}(t_1^\prime,t_2)$: fraction of problems changed from incorrect to correct;
	\item $\Delta^{c \to i}(t_1^\prime,t_2)$: fraction of problems changed from correct to incorrect.
\end{enumerate}

These metrics allow us to distinguish between improvements in initial problem-solving and the efficacy of the self-correction mechanism.

\paragraph{Baselines}
We compare MGRPO against several baselines:
\begin{itemize}
    \item \textbf{PPO (One-round)}~\citep{schulman2017proximal}: Standard Proximal Policy Optimization applied to the LLM, using outcome-based rewards. This represents a common RL alignment approach.
    \item \textbf{GRPO (One-round)}~\citep{shao2024deepseekmath}: The standard GRPO algorithm, as described in Section 3.1, without the explicit second self-correction layer. This is the direct predecessor to MGRPO.
    \item \textbf{Intrinsic Self-Correction (prompt-based)}~\citep{huang2023large}: This baseline uses the base model without further RL training and attempts self-correction purely through prompting. Accuracy@t1$^\prime$ denotes the base model's initial response accuracy under single-prompt inference, where Accuracy@t2 reflects the enhanced accuracy after executing a two-round self-correction process through iterative prompting.   
\end{itemize}
The "One-round" baselines (PPO, GRPO) do not have an explicit second turn for correction; their Acc.@t1 is their final accuracy, and Acc.@t2 metrics are not applicable (indicated by "-").

\paragraph{Implement Details}
For training, we configure the hyperparameters as follows: learning rate = 5.0e-7, LR scheduler type = cosine, warmup ratio = 0.03, beta=0.001, maximum completion length = 8196, number of generations = 8, and batch size = 32. During RL training and evaluation, model inference is performed with a maximum context length of 4096 tokens. To accelerate inference, we utilize VLLM version 0.6.3 \citep{kwon2023efficient}.

\section{Results}

The experimental results, presented in Table~\ref{tab:scaled_result}, demonstrate the effectiveness of MGRPO across all four challenging mathematical reasoning benchmarks.

Overall, MGRPO consistently outperforms both one-round PPO and one-round GRPO. For instance, on GSM8K, MGRPO achieves a final accuracy  of 95.6\%, a significant improvement over one-round GRPO's 83.4\%. This pattern of superior performance is echoed across MATH (90.4\% for MGRPO over 80.9\% for GRPO), Minerva Math (39.3\% over 35.1\%), and OlympiadBench (50.4\% over 39.9\%).

\textbf{Efficacy of the Self-correction Mechanism.} Notably, MGRPO's initial response accuracy (Acc.@t1) is identical to that of standard GRPO, as Layer 1 is the same. The subsequent improvements (Acc.@t1$^\prime$) highlight the strong contribution of the self-correction mechanism.

\textbf{Efficacy of the second turn GRPO.} The metrics $\Delta(\text{t1}^\prime,\text{t2})$, $\Delta^{i \to c}(\text{t1}^\prime,\text{t2})$, and $\Delta^{c \to i}(\text{t1}^\prime,\text{t2})$ provide deeper insights into the second turn GRPO capabilities fostered by MGRPO. The positive $\Delta(\text{t1}^\prime,\text{t2})$ values (+2.9 on MATH, +2.2 on GSM8K, +3.2 on Minerva Math, +4.9 on OlympiadBench) indicate that the explicit RL training in Layer 2 effectively hones the self-correction skill. Crucially, MGRPO demonstrates a strong ability to convert incorrect answers to correct ones, as evidenced by consistently positive $\Delta^{i \to c}(\text{t1}^\prime,\text{t2})$, which range from 2.3\% on GSM8K to 5.3\% on OlympiadBench. More importantly, it does so with minimal negative impact: the rate of changing correct answers to incorrect ones ($\Delta^{c \to i}(\text{t1}^\prime,\text{t2})$) is remarkably low across all datasets, averaging around 0.3\%. The result shows that the RL-trained self-correction is learning to apply changes judiciously, further improving the self-correction attempts.  The second turn GRPO after self-correction is not only effective but also reliable, predominantly making productive edits.

\textbf{Comparison with Intrinsic Self-Correction.} Our study also reveals that intrinsic self-correction generally fails in the absence of RL training. For baseline "Intrinsic self-correction", Acc.@t1$^\prime$ is substantially lower than Acc.@t1 in all cases, indicating that without targeted training, the model often makes its output worse when asked to reflect. The $\Delta^{c \to i}$ values are also alarmingly much higher, showing that it frequently corrupts initially correct answers. This starkly contrasts with MGRPO's performance and underscores the necessity of the previous RL framework to teach the model "how" and "when" to correct effectively. MGRPO learns to identify genuine errors and refine them, rather than indiscriminately altering its output.

\textbf{Synergistic Learning in MGRPO.} Compared to all baselines, our study suggests that the two layers of MGRPO create a synergistic learning loop. The Layer 2 GRPO, by training on self-generated correction attempts, explicitly rewards the model for identifying and fixing errors. This learned self-correction ability appears to positively influence the quality of reasoning in Layer 1 over time. As the model becomes better at self-correction, its initial reasoning chains generated in Layer 1 also improve, likely because the shared policy benefits from the enhanced understanding of error patterns and valid reasoning structures. The improved Layer 1 outputs then provide better or more easily correctable inputs for Layer 2, creating a positive feedback cycle. The selection mechanism in Layer 2, which focuses training on successful corrections, ensures that the policy updates are driven by productive refinement signals.

In essence, MGRPO’s structured approach allows the model to learn from its mistakes in a targeted manner, which provides implicit process-level supervision, leading to more robust reasoning and a significantly enhanced ability to produce accurate final answers, especially in complex, multi-step problems. An example of MGRPO inference with detailed reasoning process omitted is attached in appendix.

\section{Conclusion}
This paper introduces Multi-Layer GRPO (MGRPO), which recycles data generated during the GRPO process for self-correction learning, guiding the model to to learn to correct erroneous steps during reasoning and enabling the generative policy model to act as an implicit process reward model. Experimental results demonstrate that MGRPO achieves statistically significant improvements over GRPO. Furthermore, MGRPO can be generalized to other RL algorithms, opening new directions for dense process reward reinforcement learning in LLMs.





\section*{Impact Statement}
This paper introduces Multi-Layer GRPO (MGRPO), a method for enhancing reasoning and self-correction in large language models. The objective is to improve LLM accuracy and reliability, facilitating more dependable AI for complex tasks. This can lead to more effective AI-assisted tools in education and research  and more robust problem-solving capabilities (functional benefits).  We have not identified additional ethical issues specific to the MGRPO training approach that require distinct emphasis beyond those pertinent to general LLM advancement.

\bibliography{acl_latex}
\newpage
\appendix

\section{An example of MGRPO.}
\label{sec:appendix}

\begin{table*}[]
	\centering
	\begin{minipage}{2.0\columnwidth}\vspace{0mm}    \centering
	\caption{An example of MGRPO inference, with the detailed reasoning process omitted.} 
		\label{fig:example_MGRPO_prompt}
		\begin{sectionbox}[]
			\centering
			\small
			\begin{tabular}{p{0.97\columnwidth} c}
				\textcolor{blue}{Layer1 grpo:} \\
				\textbf{Input:}\texttt{<|begin\_of\_sentence|><|User|>Tobias is buying a new pair of shoes that costs \$95. He has been saving up his money each month for the past three months. He gets a \$5 allowance a month. He also mows lawns and shovels driveways. He charges \$15 to mow a lawn and \$7 to shovel. After buying the shoes, he has \$15 in change. If he mows 4 lawns, how many driveways did he shovel?<|Assistant|>}
				 \\\\

				\textbf{output:}\texttt{<think> First, I need to determine how much money Tobias has saved in total. The shoes cost \$95, and after purchasing them, he has \$15 left. This means he had \$110 in total savings. , I'll calculate how much he earned from mowing lawns. He mowed 4 lawns at ...Finally, I'll verify that 7 driveways yield \$49, leaving exactly \$1 left over, which matches the given information.
					\textcolor{magenta}{</think>
					Let's break down the problem step by step to find out how ...}
				}

				\\\\
				\textcolor{blue}{Layer2 grpo:} \\
				\textbf{Input:}\texttt{<|begin\_of\_sentence|><|User|>Tobias is buying a new pair of shoes that costs \$95. He has been saving up his money each month for the past three months. He gets a \$5 allowance a month. He also mows lawns and shovels driveways. He charges \$15 to mow a lawn and \$7 to shovel. After buying the shoes, he has \$15 in change. If he mows 4 lawns, how many driveways did he shovel?<|Assistant|>\textcolor{cyan}{<think> First, I need to determine how much money Tobias has saved in total. The shoes cost \$95, and after purchasing them, he has \$15 left. This means he had \$110 in total savings. , I'll calculate how much he earned from mowing lawns. He mowed 4 lawns at ...Finally, I'll verify that 7 driveways yield \$49, leaving exactly \$1 left over, which matches the given information.}\textcolor{red}{Where might I have gone wrong this time?Let me double-check carefully: }}
				\\\\

				\textbf{output:}\texttt{Total savings after earning and before buying shoes: \$110 ... Therefore, Tobias shoveled 5 driveways.</think>
					﻿
					Let's break down the problem step by step to ...	}

				\\\\
				
			\end{tabular}
		\end{sectionbox}
	\end{minipage}
\end{table*}
\end{document}